\documentclass[11pt]{article}

\usepackage[final]{acl}

\usepackage{times}
\usepackage{latexsym}

\usepackage[T1]{fontenc}

\usepackage[utf8]{inputenc}

\usepackage{microtype}

\usepackage{inconsolata}

\usepackage{graphicx}

\usepackage{amsmath}

\newcommand{\myquote}[1]{``#1''}

\title{\textsc{Annotares}: A Dataset for Extracting Logical Structures from German Statutory Texts}

\author{Ronja Schwarz \and Jannik Strötgen \\
         Karlsruhe University of Applied Sciences\\Moltkestr. 30, 76133 Karlsruhe, Germany\\
         \texttt{mail@aesz.org}\\
         \texttt{jannik.stroetgen@h-ka.de}}

\begin{document}
\maketitle
\begin{abstract}
The automatic structural analysis of legal texts is a cornerstone of legal technology, yet the extraction of their logical components remains a significant challenge.
In this paper, we introduce the task of identifying and segmenting legal conditions (\emph{Tatbestand}) and legal consequences (\emph{Rechtsfolge}) within German statutory texts.
To support this task, we present \textsc{Annotares} (Annotations of Tatbestand-Rechtsfolge Sequences), a novel dataset comprising German law texts with span-level annotations.
Spanning three distinct legal codes, the dataset is designed to evaluate both domain-specific performance and cross-statute generalizability.
We benchmark diverse architectural approaches: a rule-based baseline, CRFs, BiLSTMs, BiLSTM-CRF, and modern Transformer-based models, including BERT variants and LLM-based methods. Our results demonstrate that BERT and LLM-based models achieve superior performance in capturing the complex syntactic structures of legal language. We release our dataset to facilitate further research in automated legal reasoning.\footnote{\url{https://github.com/wilhelmines/annotares}}
\end{abstract}

\section{Introduction}
Recent advancements in natural language processing, specifically the rise of large language models (LLMs), have greatly enhanced the capacity for the nuanced structural analysis and automated reasoning required in the legal domain \cite{SiinoEtAl2025}. 
However, the precise extraction of the underlying logical structures within statutory texts -- specifically the identification of legal conditions (\emph{Tatbestand}) and legal consequences (\emph{Rechtsfolge}) within German law -- remains a significant technical challenge. This difficulty largely stems from the fact that German statutory texts are crafted for human interpretation, characterized by highly nested phrasing and implicit legal contexts that complicate automated parsing.

In Germany, the federal government maintains a manual for drafting legislation\footnote{\url{https://hdr4.bmj.de/}} to unify ministerial drafts and ensure ease of understanding, a uniform structure, and a consistent stylistic approach.
Although the first version of the manual was already released in 1991, the principles of the 2024 edition (4th ed.) have not yet been reflected in the vast majority of existing statutory texts. 
While the manual does not yet aim for machine-executable drafting for new regulations, the suggested uniform syntax and predictable structure is a prerequisite. 
In addition, recent initiatives (e.g., the Law as Code initiative of the Federal Agency for Disruptive Innovation\footnote{\url{https://www.sprind.org/en/actions/strategic-projects/law-as-code}}) propose machine-executable drafting for new regulations. Nevertheless, as of today, the overwhelming majority of existing statutory texts possess complex syntactic structures. Even with the capabilities of modern LLMs, explicitly transforming these texts into structured logical components remains a fundamental prerequisite for reliable automated legal reasoning.

Within the broader LegalNLP domain, significant progress has been made in tasks such as named entity recognition (e.g., \citealt{leitner2019fine,au-etal-2022-e}), judgment prediction (e.g., \citealt{chalkidis-etal-2019-neural,feng-etal-2022-legal}), and document summarization (e.g., \citealt{kanapala_text_2019, JainEtAl2024}).
However, these tasks usually treat legal text as a collection of entities or a sequence of labels without capturing the underlying logic defining a legal norm. 
Transforming statutory prose into a form suitable for automated reasoning requires a more fine-granular approach that segments text into its logical components: legal prerequisite conditions and the resulting legal consequences.

In this paper, we introduce a novel sequence tagging task for the legal domain to extract logical components from statutory text: the identification and segmentation of legal conditions (\emph{Tatbestand}) and legal consequences (\emph{Rechtsfolge}). For this, we introduce the \textsc{Annotares} dataset (Annotations of Tatbestand-Rechtsfolge Sequences), which spans three distinct legal codes: the Federal Data Protection Act  (\emph{Bundesdatenschutzgesetz}, BDSG), the Federal Education and Training Assistance Act (\emph{Bundesausbildungsförderungsgesetz}, BAföG), and the Federal Building Code (\emph{Baugesetzbuch}, BauGB). 

\textsc{Annotares} includes a primary corpus of more than 400 annotated sentences from the BDSG, split into training and test sets, complemented by 50 annotated sentences each from the BAföG and BauGB to serve as out-of-domain test sets for assessing cross-statute generalizability. 
The dataset contains more than 21,500 annotated tokens with approximately 43\%, 44\% and 13\% of them being annotated as legal conditions (\emph{Tatbestand}), legal consequence (\emph{Rechtsfolge}) and none, respectively. Despite the low proportion of the none class, a token-level inter-annotator agreement analysis yielded a Krippendorff's alpha of 0.89, indicating high annotation quality and robust boundary detection.

As a further main contribution of this paper, we benchmark six diverse architectural approaches: a rule-based baseline, CRFs, BiLSTMs, BiLSTM-CRF, and modern Transformer-based models, including BERT variants and LLM-based methods. 
Our results demonstrate that BERT and LLM-based models achieve superior performance, successfully capturing the complex syntactic structures inherent in legal language.

The remainder of the paper is structured as follows: After surveying related work in Section~\ref{sec:rel_work}, we define the task and describe the annotation guidelines in Section~\ref{sec:anno_guidelines}.
The dataset annotation process and detailed corpus statistics are presented in Section~\ref{sec:dataset} accompanied with detailed dataset statistics.
Finally, Section~\ref{sec:eval} describes our experimental setup and evaluation results, before we conclude and outline future research directions in Section~\ref{sec:conclusions}.

\section{Related Work}\label{sec:rel_work}

Although the legal domain is a highly specialized field, it supports an active research community.
Several works addressed legal document summarization (e.g., \citealt{kanapala_text_2019, JainEtAl2024}), legal question answering (e.g., \citealt{AskariEtAl2024, buttner-habernal-2024-answering,Dijck_Spanakis_2024}), and legal judgment prediction (e.g., \citealt{chalkidis-etal-2019-neural,feng-etal-2022-legal}). Furthermore, various works addressed text classification for various types of legal text documents (e.g., \citealt{chalkidis-etal-2021-multieurlex,wang-etal-2022-d2gclf}), including patents (e.g., \citealt{PujariEtAl2021}). 

\citet{glaser2021sentence} addressed the task of sentence boundary detection in German legal documents demonstrating that tackling NLP tasks in the legal domain requires domain adaptation similar to many other domains in other low-resource scenarios \citep{hedderich-etal-2021-survey}. 

More closely related to our work, there are also some works targeting sequence tagging tasks. Most prominently, named entity recognition (NER) in the legal domain was addressed in several languages \cite{pais-etal-2021-named,au-etal-2022-e,smadu-etal-2022-legal}. \citet{leitner2019fine,leitner-etal-2020-dataset} addressed fine-grained NER and developed a respective dataset covering German legal documents. They used Bi\-LSTMs and CRFs to perform the sequence tagging 
-- architectures that we also include in our benchmarking experiments. 
Furthermore, the development of domain-specific language models, such as LEGAL-BERT \cite{chalkidis-etal-2020-legal}, has demonstrated that pre-training on large-scale legal corpora can significantly improve performance on downstream tasks compared to general-purpose models.
In addition, research competitions such as COLIEE  \citep{goebel2024overview} have fostered innovation in legal information extraction and entailment. 

There are also recent surveys covering the LegalNLP domain. \citet{AriaiEtAl2025_survey} provide a broad overview of natural language processing in the legal domain. They discuss various tasks, existing datasets as well as used models to tackle the discussed NLP tasks in the legal domain. \citet{SiinoEtAl2025} survey recent research focussing on LLM-based approaches.

However, to the best of our knowledge, no existing work targets the extraction of logical components in statutory text. In contrast, we define the novel task of identifying and segmenting legal prerequisite conditions (\emph{Tatbestand}) and the resulting legal consequences (\emph{Rechtsfolge}). 
Consequently, the manual for drafting legislation (\textit{Handbuch der Rechtsförmlichkeit}) released by the German Federal Ministry of Justice \citep{bmj2024hdr} serves as a foundational resource for our study.
This manual provides the structural and syntactic guidelines and offers standardized examples of the uniform syntactic structures we aim to extract. Based on them, we defined our annotation guidelines for the novel task of extracting legal conditions and legal consequences in German statutory texts.

\section{Annotation Guidelines}\label{sec:anno_guidelines} 
\paragraph{Task Definition.}
We define the extraction of logical components from statutory texts as a sequence tagging task. Each token in a sentence is assigned one of three labels: legal conditions (\emph{Tatbestand}) or legal consequences (\emph{Rechtsfolge}) or as \emph{none} (the background class).

In the civil law tradition, legal conditions (\emph{Tatbestand}) and legal consequences (\emph{Rechtsfolge}) constitute the foundational structure of legal propositions. Dividing a sentence into these functional parts allows automated systems (e.g., legal knowledge graphs, retrieval-augmented generation (RAG) pipelines, and neuro-symbolic frameworks) to parse the underlying logic of a norm. By explicitly segmenting these structures, downstream applications can focus on semantic interpretation and legal reasoning without the added burden of structural disambiguation. Our approach assumes two core principles derived from the manual for drafting legislation \citep{bmj2024hdr}:  (1) \textit{Atomicity}, where each sentence contains exactly one legal statement; and (2) \textit{Completeness}, where every sentence is either a formal definition or a combination of legal conditions and legal consequences.

\paragraph{Annotation Guidelines.}
The guidelines were developed to facilitate the annotation of German statutory texts based on identifiable linguistic patterns. This ensures that even non-domain experts can perform the task in a uniform, reproducible manner.

The framework defines rules for assigning the classes (i) \emph{Tatbestand} (legal conditions), (ii) \emph{Rechtsfolge} (legal consequence), or (iii) \emph{none} (i.e., the ``outside'' class). A key structural constraint of our guidelines is the interdependency of the logical labels: if a sentence contains a \emph{Tatbestand} (legal conditions), it must also contain a \emph{Rechtsfolge} (legal consequence), and vice versa. Sentences that do not follow this conditional structure (e.g., purely introductory text) are labeled entirely as \emph{none}.

\begin{figure}
    \centering
    \includegraphics[width=\linewidth]{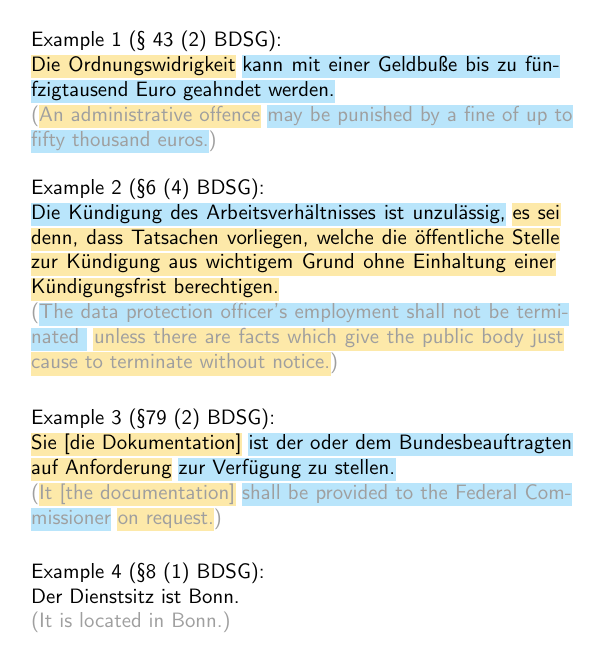}
    \caption{Representative sample sentence annotations from the BDSG. Spans identified as \emph{Tatbestand} (legal conditions) are highlighted in yellow, whiles \emph{Rechtsfolge} (legal consequence) spans are highlighted in blue. Unhighlighted text represents the \emph{none} class. English translations are sourced from the official federal version of the BDSG (\url{https://www.gesetze-im-internet.de/englisch_bdsg/englisch_bdsg.html}).}
    \label{fig:example_annotation}
\end{figure}

Representative examples of the annotation schema are depicted in Figure \ref{fig:example_annotation}.
Example 1 illustrates a standard linear structure, where the initial segment is identified as \emph{Tatbestand} (legal conditions) followed by the \emph{Rechtsfolge} (legal consequence).
Example 2 demonstrates that this logical order might be inverted, 
while Example 3 showcases a more complex syntactic structure containing multiple, interleaved spans of both conditions (\emph{Tatbestand}) and consequences (\emph{Rechtsfolge}).
Finally, Example 4 illustrates the \emph{none} class. In this instance, the sentence provides a formal definition that states a legal fact without prescribing a specific normative consequence, thus falling outside the scope of our primary logical extraction task.

\paragraph{Tatbestand.}
Legal conditions (i.e., \emph{Tatbestand}) describe the prerequisites that must be satisfied for the associated legal consequences to take effect. The manual for drafting legislation \citep{bmj2024hdr} recommends a clear structure in sentences to allow for a simple identification of the part that is the constituent element. It recommends using conditional clauses to make these elements easily identifiable. In particular, it suggests keywords such as \textit{wenn}, \textit{falls}, and \textit{sofern} (corresponding to conditionals such as ``if'', ``in case'', or ``provided that'').

However, legal conditions are not always introduced by explicit conjunctions. They may appear without a conditional clause as in Example 1, Figure \ref{fig:example_annotation}: 
\textbf{Die Ordnungswidrigkeit} [Tatbestand] kann mit einer Geldbuße [...] geahndet werden'' (\textbf{An administrative offense} [Legal Condition] may be punished by a fine [...]'').

\paragraph{Rechtsfolge.}
The legal consequence (i.e., \emph{Rechtsfolge}) specifies the normative result triggered when the corresponding legal conditions are met. In German statutory language, consequences are frequently characterized by the use of modal verbs that define the degree of obligation: \emph{kann} (permissive/discretionary), \emph{soll} (expected/guided discretion), and \emph{muss} (mandatory).

To resolve potential ambiguities, the guidelines provide specific rules for mentions of persons and institutions. When an actor is part of the situational prerequisite, they are labeled as \emph{Tatbestand} (legal conditions). If the actor is primarily the subject or object of the resulting legal action, they are included in the \emph{Rechtsfolge} (legal consequence) span.

\section{Dataset Creation and Dataset Statistics}\label{sec:dataset}

\paragraph{Annotation Procedure.}
The \textsc{Annotares} corpus was annotated by a team of six individuals, including the lead author and five additional annotators. The lead author annotated the entire corpus to provide a consistent expert baseline. Each sentence was also assigned to at least one of the other five annotators, each of whom completed a minimum of 80 sentences. This overlapping assignment strategy ensured that every sentence in the dataset had at least two independent annotations for verification.
To maintain high data quality, annotators were provided with an ``unsure'' flag for ambiguous cases. Any sentence marked with this flag was excluded from both the inter-annotator agreement calculation and the final dataset, resulting in the removal of seven sentences.

\paragraph{Data Preprocessing.}
Legal texts were sourced in XML format from the official federal repository.\footnote{\url{https://www.gesetze-im-internet.de}} 
These files contain rich metadata, including legislative years and hierarchical document positions. 
Paragraphs were extracted as individual elements, while sentence segmentation was performed using a hybrid approach: initial automated sentence splitting followed by rigorous manual correction. Manual intervention was necessary due to the characteristic length and complex punctuation of German legal prose, which often lead to errors in standard sentence splitters.

The annotation campaign utilized a custom-developed, Python-based annotation tool. While existing platforms such as Inception \citep{klie-etal-2018-inception} or Doccano \citep{doccano} offer extensive features, they often present a steep learning curve for non-experts. Our streamlined interface was designed to reduce cognitive load and simplify data management by eliminating the need for complex file-format conversions.

\paragraph{Dataset Coverage and Splits.}
\textsc{Annotares} comprises sentences from three federal German laws: the Federal Data Protection Act (\emph{Bundesdatenschutzgesetz}, \textit{BDSG}), the Federal Education and Training Assistance Act (\emph{Bundesausbildungsförderungsgesetz}, \textit{BAföG}), and the Federal Building Code (\emph{Baugesetzbuch}, \textit{BauGB}). The \textit{BDSG} was annotated in its entirety and serves as the primary corpus for training and evaluation. It was randomly split into training, validation, and test sets containing 351, 43, and 45 sentences, respectively. The \textit{BAföG} and \textit{BauGB} were sampled (50 sentences each) to serve as distinct out-of-domain test sets to evaluate cross-statute generalizability. 
These samples were selected based on structural parity, specifically matching the word count and frequency distributions observed in the primary \textit{BDSG} corpus.

\paragraph{Dataset Statistics.}
As shown in Table~\ref{tab:corpus_statistics}, the three statutes exhibit similar structural patterns, with mean sentence lengths ranging from 38.1 to 42.4 tokens. The \textit{BDSG} displays a higher proportion of the \textit{none} class (+8 percentage points) compared to the sampled sets. This discrepancy is likely due to the inclusion of the complete \textit{BDSG} text, which encompasses introductory and administrative clauses that lack a conditional structure, whereas the sampled sets targeted more dense normative content.

\begin{table*}
  \centering
  \begin{tabular}{ll c c c c c}
    \hline
     &  &  &  & \multicolumn{3}{c}{Coverage of tokens (\%)} \\
     & & & & Rechtsfolge & Tatbestand & None \\
     Law& Splits& Sentences& Tokens& (leg.\,consequence) & (leg.\,conditions) & (outside) \\
    \hline
    BDSG & Train, Val, Test & 439 & 17,524 & 44.3 & 40.6 & 15.1 \\
    BAföG & Test & \phantom{0}50 & \phantom{0}2,121 & 38.0 & 57.6 & \phantom{0}4.4 \\
    BauGB & Test & \phantom{0}50 & \phantom{0}1,905 & 46.3 & 46.6 & \phantom{0}7.1 \\
    
    \hline
    overall & & 539 & 21,550 & 43.8 & 42.8 & 13.4 \\
    \hline
  \end{tabular}
  \caption{Statistics for the three splits of the \textsc{Annotares} dataset.}
  \label{tab:corpus_statistics}
\end{table*}

\paragraph{Linguistic Preprocessing.}
To provide sophisticated features for traditional machine learning models (e.g., CRFs), we enriched the dataset with dependency parsing information in the CoNLL-U format \citep{nivre-etal-2016-universal}.  
For this, we evaluated two parsers: spaCy (\texttt{de\_core\_news\_sm}) \citep{honnibal_2020_spacy} and Stanza (\texttt{UD-German\_GSD}) \citep{qi-etal-2020-stanza}.
Following a preliminary comparison in which Stanza demonstrated a slight performance lead, we selected the Stanza model to generate the dependency features used in our benchmarking experiments.

\paragraph{Inter-Annotator Agreement.}
We employed Krippendorff's Alpha ($\alpha$) to measure agreement, as it robustly handles nominal labels and varying annotation counts per token. In our setup, the lead author (Annotator 00) served as the expert reference, having annotated the entire corpus. Individual agreement scores were calculated between the expert and each contributing annotator. The datasets from the \textit{BAföG} and the \textit{BauGB} were annotated by two persons each, while the \textit{BDSG} includes the results of six different annotators.

As shown in Table~\ref{tab:krippendorff_details}, the datasets achieved high reliability scores: $\alpha$=0.897 for \textit{BAföG}, $\alpha$=0.926 for \textit{BauGB}, and a cumulative $\alpha$=0.893 for the \textit{BDSG}. Following the initial campaign, an adjudication step was performed by the lead author to harmonize the final labels. In 84.4\% of cases, the annotations were either identical or resolved through deterministic corrections of trivial errors (e.g., adjusting misplaced punctuation at span boundaries). The remaining disputes were resolved based on the expert's deeper task understanding, with the final labels favoring the expert in 12.7\% of instances and the contributing annotator in 2.9\%.

\begin{table}
  \centering
  \begin{tabular}{lcc}
    \hline
    Dataset & Annotators & $\alpha$ \\
    \hline
    BAföG & All (2) & 0.897 \\
    \hline
    BauGB & All (2) & 0.926 \\
    \hline
    BDSG & All (6) & 0.893 \\
        & 00 \& 01 & 0.896 \\
         & 00 \& 02 & 0.858 \\
         & 00 \& 03 & 0.826 \\
         & 00 \& 04 & 0.942 \\
         & 00 \& 05 & 0.923 \\
    \hline
  \end{tabular}
  \caption{Inter-annotator agreement analysis on the \textsc{Annotares} corpus based on Krippendorff's $\alpha$. Individual agreement scores between the expert and each contributing annotator per source.}
  \label{tab:krippendorff_details}
\end{table}

\section{Experimental Setup}

We evaluate six distinct architectural approaches for the logical segmentation of statutory text. These models were selected to represent a spectrum of complexity, computational requirements, and interpretability.

As a baseline for simplicity and human interpretability, we implemented a \textbf{rule-based approach}.
We further included a \textbf{Conditional Random Field (CRF)} model as a representative for a probabilistic model \citep{Lafferty2001ConditionalRF} to represent traditional probabilistic sequence modeling, which has historically been the standard for linguistic structure extraction.

To capture long-range dependencies, we implemented \textbf{Bi-directional Long Short-Term Memory (BiLSTM)} networks and a combined \textbf{BiLSTM-CRF} architecture. The latter has consistently demonstrated high performance in named entity recognition and related sequence tagging tasks \citet{lample-etal-2016-neural}.
 
We further benchmarked several variants of the \textbf{BERT} architecture \cite{devlin-etal-2019-bert}. Our selection includes general-purpose models such as \emph{mBERT} and \emph{DistilBERT} \citep{Sanh2019DistilBERTAD}, as well as domain-specific variants like \emph{LEGAL-BERT} \cite{chalkidis-etal-2020-legal} and \emph{BERT-LER}. These models utilize the Transformer's encoder block to generate nuanced, context-aware representations \cite{NIPS2017_3f5ee243}.

Finally, we evaluated the zero-shot and few-shot capabilities of modern \textbf{Large Language Models (LLMs)}, specifically \emph{Claude 3 Haiku} and \emph{Claude 3 Opus} \citep{anthropic2024claude3}. Unlike the encoder-only BERT models, these are decoder-only architectures \cite{radford2018improving} capable of following complex instructions through natural language prompts.

Detailed hyperparameter configurations for all evaluated models are provided in the appendix (Appendix A).

\paragraph{Rules.}
We implemented a sequential three-stage rule-based system. The first rule matches verb-first (V1) sentence structures, which is a common indicator of conditional clauses in German legal prose.
The second rule utilizes a set of keywords (e.g., \emph{wenn}, \emph{falls}) as indicators for a \emph{Tatbestand} (legal conditions). These sentences are split at the primary comma, with the remaining segment labeled as \emph{Rechtsfolge} (legal consequence). 
Finally, a catch-all rule assigns the \emph{none} label to any sentence not captured by the preceding rules.

\paragraph{CRF.}

The CRF was implemented using the \texttt{python-crfsuite} library with default hyperparameters. Feature engineering included token-level attributes (text, casing, numerical flags), part-of-speech (POS) tags, and dependency labels. To capture local context, we utilized n-grams and a sliding window that incorporates the POS and dependency tags for the immediate neighbors tokens.

\paragraph{BiLSTM and BiLSTM-CRF.}
Both neural architectures utilize pre-trained \texttt{fastText} embeddings \citep{grave-etal-2018-learning} trained on German Common Crawl and Wikipedia data. These embeddings are concatenated with learned embeddings for POS and dependency tags before being processed by a Bi-directional LSTM. In the standard BiLSTM, a linear layer with a softmax activation provides the class probabilities. The BiLSTM-CRF variant extends this by passing the LSTM outputs into a CRF layer to model label dependencies. Both models were implemented using \texttt{PyTorch}.

\paragraph{BERT Variants.}
We evaluated twelve German-centric and multilingual models sourced from the Hugging Face Hub. Following a preliminary screening, the four top-performing models were selected for full evaluation: \textit{BERT-LER}, \textit{DistilBERT}, \textit{LEGAL-BERT}, and \textit{mBERT}.\footnote{mrm8488/bert-base-german-finetuned-ler (BERT-LER), dbmdz/distilbert-base-german-europeana-cased (DistilBERT), nlpaueb/legal-bert-base-uncased (LEGAL-BERT) , google-bert/bert-base-multilingual-cased (mBERT).} Our architecture concatenates the BERT hidden states with POS and dependency tag embeddings, followed by a linear classification layer. Models were trained across three random seeds to ensure statistical robustness.

\paragraph{Large Language Models (LLMs).}
We evaluated \textit{Claude Opus 4.6} and \textit{Claude Haiku 4.5} using two distinct prompting strategies. 
The first prompt was manually created with general prompt suggestions provided by ChatGPT, while the second prompt was created by Claude Sonnet 4.5 with the goal to adapt it for Claude Opus.
Prompt 1 serves as a textual representation of the annotation guidelines, including class definitions, mandatory structural rules, and few-shot examples. Prompt 2 is an optimized version designed to leverage the models' reasoning capabilities. It incorporates structured POS and dependency tag mappings and extended examples to guide the segmentation process. Both models were tested in zero-shot and few-shot configurations to assess their inherent understanding of legal logic. The full text of both prompts is available in Appendix B.

\section{Evaluation}\label{sec:eval}
In this section, we provide a comprehensive evaluation of the automated models against the human-annotated gold standard. We structure our analysis into four parts: first, we assess in-domain token-level performance on the primary BDSG test split. We then conduct a detailed error analysis and evaluate precise boundary detection using exact span matches. Third, we test the models' cross-statute generalization capabilities on the out-of-domain BAföG and BauGB test sets. Finally, we present an ablation study to quantify the contribution of structural linguistic features, specifically part-of-speech and dependency tags, to the overall performance.

All methods were evaluated across the three \textsc{Annotares} test sets. We report Macro F1 score ($mF1$) as our primary metric to account for the uneven distribution of tokens across the three classes (cf.\ Table \ref{tab:corpus_statistics}), alongside accuracy for completeness.

\begin{table}
    \centering
    \begin{tabular}{lcc}
        \hline
        Method & Accuracy & mF1 \\
        \hline
        Rules & 0.161 & 0.197 \\
        \hline
        CRF & 0.711 & 0.608 \\
        BiLSTM & 0.529 & 0.357 \\
        BiLSTM + CRF & 0.572 & 0.480 \\
        \hline
        BERT-LER & \underline{0.842} & \underline{0.782} \\
        DistilBERT & 0.784 & 0.713 \\
        LEGAL-BERT & 0.787 & 0.740 \\
        mBERT & \textbf{0.854} & \textbf{0.798} \\
        \hline
        LLM Prompt 1 + Haiku & 0.826 & 0.713 \\
        LLM Prompt 1 + Opus & 0.820 & 0.735 \\
        LLM Prompt 2 + Haiku & 0.703 & 0.606 \\
        LLM Prompt 2 + Opus & 0.678 & 0.619 \\
        \hline
    \end{tabular}
    \caption{Comparison of the approaches on the BDSG test split of the \textsc{Annotares} data set. Best scores in bold, second best underlined.}
    \label{tab:result_comp}
\end{table}

\paragraph{In-Domain Experiments.}
Table \ref{tab:result_comp} summarizes the performance of all evaluated methods on the \emph{BDSG} test split.
The results highlight a significant performance gap between traditional baselines and modern Transformer-based architectures.
The manual rule-based system performs poorly ($mF1=0.197$), largely due to its inability to handle the syntactic complexity of legal prose, resulting in an over-reliance on the \emph{none} class  (see Figure \ref{fig:conf_rules_crf} in the appendix).
Similarly, the BiLSTM architectures struggled ($mF1<0.50$), likely due to the limited size of the training data and no applied transfer learning.

All other methods show promising results for the novel task of identifying and segmenting legal conditions (\emph{Tatbestand}) and legal consequences (\emph{Rechtsfolge}) within German statutory texts. 
mBERT achieved the highest overall performance with an accuracy of $85.4\%$ and an $mF1$ score of $0.798$.
This suggests that multilingual pre-training provides a robust foundation for capturing the nuances of German legal logic.
LLMs (Haiku and Opus) utilizing the guideline-based prompt 1 demonstrated competitive results, nearly matching their performance of domain-specific models like \emph{BERT-LER}.
However, prompt 2 (despite incorporating explicit linguistic features such as dependency tags) yielded lower $mF1$ scores.
This suggests that increasing prompt complexity may introduce noise that detracts from the model's instruction-following capabilities.

\paragraph{Error Analysis and Span Matching.}
Confusion matrices (figures \ref{fig:conf_rules_crf} to \ref{fig:conf_llm} in the appendix) for all approaches reveal a systematic bias: most models favor the \emph{Rechtsfolge} (legal consequence) over the other classes.
Most models exhibit a consistent performance hierarchy where \emph{Rechtsfolge} (legal consequence) is most accurately identified, followed by \emph{Tatbestand} (legal conditions), and finally the \emph{none} class.
A notable exception is the BiLSTM model, which failed to predict the \emph{none} class entirely.

To gain deeper insights, we evaluated exact span matches (Table \ref{tab:span_match}). 
A span is only considered correct if its boundaries match the gold standard exactly without gaps or shifts.
We specifically highlight performance on the \emph{none} class, as it serves as a proxy for how well a model internalized the underlying structural rules of the text (e.g., distinguishing functional legal rules from background definitions). 
While the BERT-based models remained consistent ($0.5\pm0.1$ match rate for the \emph{none} class), \emph{Opus} with Prompt 2 exhibited a unique error profile: it reversed the common class-priority and achieved the highest exact span match for the \emph{none} class ($0.875$), despite lower overall token accuracy.

The discrepancy between token accuracy and lower span-match scores for neural models underscores the difficulty of the task. 
This is most dramatically illustrated by the BiLSTM-CRF, which achieved a 0.772 token accuracy for \emph{Rechtsfolge} but completely failed to predict a single exact span for \emph{Rechtsfolge} and \emph{Tatbestand} (0.000).
Even if a model correctly identifies the core of a legal condition, failing to precisely identify the boundary results in a total span failure.
Finally, with the notable exception of the \emph{Opus} model (in particular with Prompt 2), we observed that the systems generally struggled to detect or enforce the underlying logical rule of the corpus, i.e., the fundamental incompatibility of \emph{none} labels with \emph{Tatbestand} (legal conditions) / \emph{Rechtsfolge} (legal consequence) within the same normative sentence.

\begin{table*}
    \centering
    \begin{tabular}{lcccccc}
        \hline
        & \multicolumn{3}{c}{Token Accuracy} & \multicolumn{3}{c}{Exact Span Match} \\
        \cline{2-4} \cline{5-7}
        Method & Rechtsfolge & Tatbestand & None & Rechtsfolge & Tatbestand & None \\
        \hline
        Rules & 0.029 & 0.184 & 1.000 & 0.061 & 0.047 & 1.000 \\
        CRF & 0.770 & 0.675 & 0.449 & 0.367 & 0.465 & 0.500 \\
        BiLSTM & 0.706 & 0.368 & 0.000 & 0.042 & 0.032 & 0.000 \\
        BiLSTM + CRF & 0.772 & 0.408 & 0.087 & 0.000 & 0.000 & 0.042 \\
        BERT-LER & 0.893 & 0.793 & 0.542 & 0.389 & 0.595 & 0.458 \\
        DistilBERT & 0.885 & 0.704 & 0.504 & 0.306 & 0.444 & 0.500 \\
        LEGAL-BERT & 0.860 & 0.719 & 0.509 & 0.368 & 0.460 & 0.417 \\
        mBERT & 0.876 & 0.850 & 0.589 & 0.438 & 0.579 & 0.542 \\
        LLM Prompt 1 + Haiku & 0.903 & 0.796 & 0.409 & 0.408 & 0.395 & 0.375 \\
        LLM Prompt 1 + Opus & 0.861 & 0.786 & 0.676 & 0.469 & 0.581 & 0.750 \\
        LLM Prompt 2 + Haiku & 0.760 & 0.652 & 0.523 & 0.163 & 0.209 & 0.625 \\
        LLM Prompt 2 + Opus & 0.662 & 0.683 & 0.761 & 0.306 & 0.186 & 0.875 \\
        \hline
    \end{tabular}
    \caption{Token accuracy and exact span match per class (BDSG).}
    \label{tab:span_match}
\end{table*}

\paragraph{Cross-Statute Generalization.}

\begin{table*}
    \centering
    \begin{tabular}{lcccccccccc}
        \hline
            & \multicolumn{2}{c}{Rules}
            & \multicolumn{2}{c}{CRF}
            & \multicolumn{2}{c}{BiLSTM}
            & \multicolumn{2}{c}{BiLSTM + CRF}
            & \multicolumn{2}{c}{mBERT} \\
            & Acc & mF1 & Acc & mF1 & Acc & mF1 & Acc & mF1 & Acc & mF1 \\
        \hline
        BAföG & 0.231 & 0.235 & 0.595 & 0.465 & 0.470 & 0.320 & 0.517 & 0.566 & 0.854 & 0.868 \\
        BauGB & 0.315 & 0.313 & 0.685 & 0.648 & 0.490 & 0.338 & 0.455 & 0.455 & 0.843 & 0.812 \\
        BDSG  & 0.161 & 0.197 & 0.711 & 0.608 & 0.529 & 0.357 & 0.572 & 0.480 & 0.854 & 0.798 \\
        \hline
    \end{tabular}
    \caption{Results of the cross-statute generalization study showing the scores on the three \textsc{Annotares} test sets.}
    \label{tab:additional_test_sets}
\end{table*}

To evaluate the robustness of the models across different legal domains, we conducted a generalization study using the \emph{BAföG} and \emph{BauGB} test sets. For this experiment, we utilized the best-performing configurations from our in-domain tests: the \emph{mBERT} model and the top-performing seeds for the BiLSTM variants. We excluded the LLMs from this specific comparison to focus on the transferability of the trained sequence taggers. 

As shown in Table \ref{tab:additional_test_sets}, the rule-based baseline demonstrated slightly higher performance on the \emph{BAföG} and \emph{BauGB} test sets than on the \emph{BDSG}. This suggests that the sampled sentences from these statutes happen to align more closely with the specific V1 and keyword-based structures defined in our rules, though overall performance remains insufficient for practical use.

The CRF yielded mixed results. While it achieved its highest accuracy on the in-domain \textit{BDSG} set ($0.711$), its $mF1$ score was significantly higher on the \textit{BauGB} set ($0.648$). The BiLSTM variants showed a slight in-domain bias, with accuracy drops of up to $0.059$ when moving to out-of-domain data. However, the BiLSTM-CRF showed inconsistent generalization behavior, achieving its highest $mF1$ on the \textit{BAföG} set ($0.566$) despite a lower accuracy, indicating a high variance in how the CRF layer handles the label transitions in different legislative contexts.

The mBERT model exhibited the most impressive generalization capabilities. It maintained a near-constant accuracy across all three statutes (ranging from $0.843$ to $0.854$). Remarkably, the $mF1$ scores for the out-of-domain sets (\textit{BAföG}: $0.868$; \textit{BauGB}: $0.812$) actually exceeded the in-domain score ($0.798$). This suggests that the model has successfully captured universal structural patterns of German statutory law that transcend specific administrative domains.

The performance fluctuations across the more traditional models (CRF and BiLSTM) further reinforce our earlier observation: a training set of 351 sentences appears insufficient for these architectures to reach a stable state of generalization. In contrast, the Transformer-based approach leverages its extensive pre-training to overcome these data limitations, proving highly effective for cross-statute logical segmentation.

\paragraph{Ablation Study.}

\begin{table}
    \centering
    \begin{tabular}{lccc}
        \hline
        Model & POS/DEP & Acc & mF1\\
        \hline     
        CRF & yes & 0.711 & 0.608 \\
        CRF & no & 0.691 & 0.589 \\
        BiLSTM & yes & 0.529 & 0.357 \\
        BiLSTM & no & 0.534 & 0.259 \\
        BiLSTM + CRF & yes & 0.572 & 0.480 \\
        BiLSTM + CRF & no & 0.582 & 0.292 \\
        mBERT & yes & 0.854 & 0.798 \\
        mBERT & no & 0.757 & 0.679 \\
        \hline
    \end{tabular}
    \caption{Results of the ablation study.}
    \label{tab:ablation_study}
\end{table}

To quantify the contribution of structural linguistic information to the segmentation task, we conducted an ablation study by removing part-of-speech (POS) and dependency (DEP) tags from our top-performing models. All ablated models were evaluated on the \textit{BDSG} test set using the same seed and architecture configurations as the primary experiments. Note that our ablated BiLSTM-CRF follows the standard architecture proposed by \citet{lample-etal-2016-neural}, utilizing only token-level embeddings.

As shown in Table~\ref{tab:ablation_study}, the inclusion of POS and DEP features significantly enhances performance across nearly all architectures, particularly in terms of $mF1$. The Transformer (mBERT) model exhibited the most dramatic sensitivity to these features. Removing linguistic tags caused the $mF1$ to drop from $0.798$ to $0.679$ and accuracy to fall by nearly 10 percentage points. This indicates that while mBERT has strong internal representations, explicit syntactic features are crucial for anchoring legal logic to specific grammatical structures.

The BiLSTM variants presented a unique behavior. While their accuracy slightly increased or remained stable upon removing linguistic features, their $mF1$ scores dropped severely (e.g., from $0.480$ to $0.292$ for BiLSTM-CRF) -- highlighting that token accuracy should not be used as the only metric for this task. An inspection of the confusion matrices (figures 6 to 8 in the appendix) reveals that without linguistic context, these models essentially collapse toward the majority class (\textit{Rechtsfolge}). The ablated BiLSTM, for instance, showed no awareness of the \textit{none} class and lost significant precision in identifying \textit{Tatbestand} boundaries.

Similarly, the CRF model's performance decreased across both metrics, though it remained more robust than the neural variants. Across all models, the removal of POS and DEP tags exacerbated the systemic bias toward the majority class. This suggests that the structural language information we provide serves as a necessary counterweight to the statistical bias inherent in the dataset, allowing models to identify the logical conditions (\textit{Tatbestand}) that are often defined by specific syntactic dependencies rather than just vocabulary.

\section{Conclusions and Future Work}\label{sec:conclusions}
In this paper, we introduced \textsc{Annotares}, a novel dataset for the automated logical segmentation of German statutory text into \textit{Tatbestand} (conditions) and \textit{Rechtsfolge} (consequences). Our benchmarking of six architectural approaches demonstrates that while traditional and neural baselines struggle with data scarcity, Transformer-based models (specifically mBERT) provide a robust foundation for this task. Notably, although LLMs showed lower overall token-level performance, they outperformed all other models 
in exact span matching for the background (\textit{none}) class.

The results confirm that structural linguistic features (POS and dependency tags) are essential for capturing legal logic in traditional and fine-tuned neural architectures, as their removal led to a significant performance collapse. In contrast, explicitly injecting these structural features into LLM prompts (Prompt 2) degraded overall token-level performance, suggesting that the added prompt complexity introduces noise rather than aiding extraction.

Building on the \textsc{Annotares} foundation, future work should focus on:
(i) {Dataset Expansion:} Incorporating EU and DACH-region statutes to enhance cross-jurisdictional generalizability;
(ii) {Two-Stage Modeling:} Developing pipeline architectures that first filter normative vs.\ non-normative sentences before performing token-level extraction; and
(iii) {Domain-Specific Tooling:} Evaluating the performance of dependency parsers fine-tuned specifically on legal corpora to replace general-domain baselines.

By providing \textsc{Annotares} and this initial benchmark, we aim to facilitate new research into logic-aware NLP for the legal domain.

\section*{Acknowledgments}
The authors want to thank Ben Dondelinger, Frederik Höft, Hannah Henze, Katharina Wittemann, Marius Höll, Markus Roos and Tanja Schwarz for their support during the annotation campaign. In addition, we thank the anonymous reviewers for their helpful comments to further improve the paper.

\paragraph{Generative AI Assistance Declaration.} During the preparation of this work, the authors used generative AI tools to rephrase and proofread text content. After using these tools, the authors reviewed and edited the content as needed and take full responsibility for the publication's content.

\bibliography{custom,anthology-1,anthology-2}


\appendix

\section{Hyperparameter}

No extensive hyperparameter search has been carried out. The CRF approach uses the default settings by the pycrfsuite package. The features provided to the model are the word/token in lower case form, the ending of previous word, the beginning of following word, is it completely capitalized, is the first letter capitalized, is it a digit, its POS- and dependency-tag. For the neighboring words the features are the lower case word/token, is the first letter capitalized, its POS- and dependency-tag.

The BiLSTM and BiLSTM-CRF models are based on the paper by \citet{lample-etal-2016-neural}. The implementation uses the pytorch library. From seeds 1-3 seed 3 was selected. The BiLSTM was trained for 11 epochs and the BiLSTM+CRF for 37 epochs. The epochs were selected by a patience of 10 and a max epoch count of 100. The initial learning rate is 0.01 with an additional learning rate decay of 0.05 using the formula:

$$\text{Current learning rate} = \frac{0.01}{1 + (0.05 \times \text{epoch})}$$

The selected optimizer is SGD and a clip of 5 was used. The pretrained word embeddings are cc.de.300.vec \citep{grave-etal-2018-learning} with a word embedding dimension of 100. The model uses a bidirectional LSTM with 100 hidden dimensions and a bidirectional LSTM with 25 hidden dimensions for the char embeddings. Two additional embedding layers are used for the POS- and dependency-tags with 25 dimensions each. A dropout of 0.5 is applied between the embeddings and the final LSTM layer. The classification is done with a linear classification layer. The CRF variant extends the BiLSTM with a CRF layer using the torchcrf package.

The transformer approach combines a pretrained BERT model with two custom embeddings for POS- and dependency-tags. The model was trained for 20 epochs with a patience of 10 and a max epoch count of 100. Based on hardware constraints the batch size of 8 was selected. The seeds 1-3 were trained and seed 1 was selected. After the concatenation of the three embedding layers a dropout of 0.1 was applied. The dropout layer feeds into a linear layer for the classification step. The optimizer is AdamW. The custom embeddings have 64 dimensions each and an initial learning rate of 0.001. The learning rate for the BERT model starts with 0.00002. The scheduler ReduceLROnPlateau was chosen for the learning rate. The settings are mode max, factor 0.5 and patience 2.

\section{Prompts}

The Claude Opus 4.6 as the frontier model was chosen based on the LMArena scores in February 2026. To avoid changes in the scripts between the models, a comparison against a lower tier model was done with Claude Haiku-4.5-20251001. The API was inferred with temperature set to 0 and a batch size of 3-5 sentences. Two different prompts were used. The first prompt was manually crafted on the 20. 03. 2026 with phrasing suggestions by ChatGPT. The second prompt was created by Claude Sonnet 4.5 Extended with access to the annotation guidelines, the BDSG training data and the manually crafted prompt. The chat bot was tasked to tailor it to Claude Opus 4.6. Claude Sonnet was allowed to make changes to the inference script. 

\subsection*{Prompt 1} 

\textbf{System Prompt}

You are an expert in German syntax and legal NLP.
Your task is the correct token classification of legal conditions and consequences.

ANNOTATION RULES (binding):
Classify EVERY single token into one of the following categories:

\begin{itemize}
    \item 2 (Tatbestand) = Condition/Prerequisite (When? Under what conditions? For whom? Scope of application? Actor/Subject?)
    \item 1 (Rechtsfolge) = Legal Consequence/Order (What applies/must be done? Who must/should/can? How to implement? The specific legal action/mandate.)
    \item 3 (None) = Neither.
\end{itemize}

Decision Rules \& Syntactic Patterns:

\begin{enumerate}
    \item Co-occurrence Requirement: If NO clearly recognizable condition AND consequence appear TOGETHER in the sentence, ALL tokens must be classified as 3. Only if BOTH are present may tokens be marked as 1 and 2.
    \item Actors and Subjects as Conditions: In sentences dictating an obligation or rule, the subject defining the actor (e.g., \myquote{Der Verantwortliche}, \myquote{Die oder der Gewählte}) or the scope of application (e.g., \myquote{Für Verarbeitungen\ldots}) acts as the condition and must be classified as 2. The predicate dictating the action (e.g., \myquote{hat \ldots zu unterrichten}) is the consequence (1).
    \item Triggers for Conditions (Class 2):
    \begin{itemize}
        \item Conjunctions: \myquote{wenn}, \myquote{sofern}, \myquote{soweit}, \myquote{falls}, \myquote{im Falle}.
        \item V1-Syntax (Verb-First): Clauses starting with the finite verb (e.g., \myquote{Stellt die Behörde fest, so\ldots}) function as conditions (2), often followed by the consequence (1) triggered by \myquote{so}.
        \item Restrictive Relative Clauses: Clauses defining the affected persons or scope (e.g., \myquote{wer wissentlich\ldots}, \myquote{Profiling, das zur Folge hat\ldots}) act as conditions (2).
    \end{itemize}
    \item Structural Markers: Paragraph numbers (e.g., \myquote{( 1 )}) and enumeration items (e.g., \myquote{ 1.}, \myquote{a)}) inherit the class of the text span they introduce.
    \item Discontinuous Spans: Spans can interrupt each other. A consequence (1) can begin, be interrupted by a conditional clause (2), and then continue (1).
\end{enumerate}

Transformation Aid:
If the sentence seems to contain only a consequence, mentally test: \myquote{Wenn <Tatbestand/Akteur>, dann <Rechtsfolge>.}
If no plausible Tatbestand (including an actor or scope) can be reconstructed, mark everything as 3.

EXAMPLE 1 (Actor as Condition):

Input:  {"0": "Der", "1": "Verantwortliche", "2": "hat", "3": "die",
         "4": "betroffene", "5": "Person", "6": "zu", "7": "unterrichten", "8": "."}
Output: {"0": 2, "1": 2, "2": 1, "3": 1, "4": 1, "5": 1, "6": 1, "7": 1, "8": 1}

EXAMPLE 2 (Classical Condition):

Input:  {"0": "Mit", "1": "dem", "2": "Ausscheiden", "3": "aus", "4": "dem",
         "5": "Amt", "6": "als", "7": "Leiterin", "8": "oder", "9": "Leiter",
         "10": "der", "11": "Aufsichtsbehörde", "12": "eines", "13": "Landes",
         "14": "endet", "15": "zugleich", "16": "die", "17": "Funktion",
         "18": "als", "19": "Stellvertreter", "20": "."}
Output: {"0": 2, "1": 2, "2": 2, "3": 2, "4": 2, "5": 2, "6": 2, "7": 2,
         "8": 2, "9": 2, "10": 2, "11": 2, "12": 2, "13": 2, "14": 1,
         "15": 1, "16": 1, "17": 1, "18": 1, "19": 1, "20": 1}

OUTPUT FORMAT:
You MUST return exclusively a JSON object. The object must contain a key \texttt{results}, which is a list of dictionaries. Each dictionary maps the numeric index of the token (as a string) to the integer value of the class (1, 2, or 3). NO explanations, NO Markdown.

\textbf{User Prompt Template}

Classify these sentences token by token:

Sentence 0: <JSON-encoded indexed token dictionary for sentence 0>

Sentence 1: <JSON-encoded indexed token dictionary for sentence 1>

\ldots

Expected pure JSON format (example output for 2 sentences):
{
  "results": [
    {"0": 2, "1": 2, "2": 3, "3": 1, "4": 1},
    {"0": 3, "1": 3, "2": 3}
  ]
}

\subsection*{Prompt 2}

\textbf{System Prompt}

You are an expert in German syntax, dependency grammar, and legal NLP.
Your task is token classification of German legal sentences into three classes
using BOTH the surface text AND the syntactic annotations provided.

CLASSES:

\begin{itemize}
    \item 2 (Tatbestand) = Condition / Prerequisite.
    Answers: When? Under what circumstances? For whom?
    Who is the actor/subject whose existence triggers the rule?
    What is the scope of application?
    \item 1 (Rechtsfolge) = Legal Consequence / Obligation.
    Answers: What must be done? Who must/may/shall act?
    What is the specific legal mandate or action?
    \item 3 (None) = Neither condition nor consequence.
\end{itemize}

FUNDAMENTAL CONSTRAINT --- Co-occurrence Rule:
Assign classes 1 and 2 ONLY when BOTH a recognisable condition AND a recognisable
consequence are present in the same sentence. If you cannot identify BOTH, mark every
token as 3. Apply the mental test: \myquote{Wenn <Tatbestand>, dann <Rechtsfolge>.}
If no plausible Tatbestand can be reconstructed from the sentence itself, mark everything 3.

SYNTACTIC DECISION RULES (use the dep/pos/head fields):

\begin{enumerate}
    \item Conditional adverbial clause (\texttt{dep=advcl} + conditional marker):
    A clause attached via \texttt{dep=advcl} whose subordinator (\texttt{dep=mark}) is one of
    \myquote{wenn}, \myquote{sofern}, \myquote{soweit}, \myquote{falls}, \myquote{im Falle} $\rightarrow$ entire clause = class 2.
    The main clause (\texttt{dep=root} and its dependents) $\rightarrow$ class 1.

    \item V1 conditional (verb-first clause, \texttt{dep=advcl}, no SCONJ marker):
    Finite verb first in a \texttt{dep=advcl} clause (e.g., \myquote{Stellt X fest, so \ldots}) $\rightarrow$ class 2.
    The particle \myquote{so} (\texttt{dep=advmod} on root) signals the transition to class 1.

    \item Actor-as-condition (\texttt{nsubj} of obligation predicate):
    \myquote{<Actor> hat/ist/muss/soll <zu+Infinitiv>} with no other explicit condition:
    \begin{itemize}
        \item \texttt{dep=nsubj} (+ its subtree) $\rightarrow$ class 2.
        \item Root predicate and its dependents $\rightarrow$ class 1.
    \end{itemize}
    EXCEPTION: if another explicit condition exists (\texttt{dep=advcl} with KOUS, or a
    demonstrative phrase like \myquote{in diesem Fall}), that is class 2; subject $\rightarrow$ class 1.

    \item Demonstrative conditions: PP with PDAT (e.g., \myquote{in diesem Fall}) $\rightarrow$ class 2.

    \item Discontinuity is allowed. A class-1 span may be interrupted by class-2 tokens.

    \item Structural tokens (paragraph markers, enum items, punctuation) inherit
    the class of the span/governor they belong to.

    \item Pure definitions and cross-references $\rightarrow$ class 3 for every token.
\end{enumerate}

POS CHEAT-SHEET (STTS):
VVFIN/VAFIN = finite verb, VVINF/VVIZU = infinitive, VVPP = participle,
PTKVZ = separable prefix, KOUS = subordinating conjunction,
KON = coordinating conjunction, NN = noun, PPER = pronoun, ART = article,
PDAT/PDS = demonstrative.

DEP CHEAT-SHEET:
\texttt{root}, \texttt{nsubj}, \texttt{obj}, \texttt{obl}, \texttt{advcl}, \texttt{acl},
\texttt{ccomp}, \texttt{xcomp}, \texttt{mark}, \texttt{cop}, \texttt{aux},
\texttt{advmod}, \texttt{compound:prt}, \texttt{det}, \texttt{case}, \texttt{punct}.

WORKED EXAMPLES:

Example 1 --- V1-conditional $\rightarrow$ class 2, then consequence $\rightarrow$ class 1:

Sentence: "Stellt die Aufsichtsbehörde einen Verstoß fest , so ist sie befugt ."
Input:
{"sent": "Stellt die Aufsichtsbehörde einen Verstoß fest , so ist sie befugt .",
 "tokens": {
   "0":  {"w":"Stellt",           "pos":"VVFIN", "dep":"advcl",       "hd":9},
   "1":  {"w":"die",              "pos":"ART",   "dep":"det",          "hd":2},
   "2":  {"w":"Aufsichtsbehörde", "pos":"NN",    "dep":"nsubj",        "hd":0},
   "3":  {"w":"einen",            "pos":"ART",   "dep":"det",          "hd":4},
   "4":  {"w":"Verstoß",          "pos":"NN",    "dep":"obj",          "hd":0},
   "5":  {"w":"fest",             "pos":"PTKVZ", "dep":"compound:prt", "hd":0},
   "6":  {"w":",",                "pos":"\$,",    "dep":"punct",        "hd":0},
   "7":  {"w":"so",               "pos":"ADV",   "dep":"advmod",       "hd":9},
   "8":  {"w":"ist",              "pos":"VAFIN", "dep":"cop",          "hd":9},
   "9":  {"w":"sie",              "pos":"PPER",  "dep":"nsubj",       "hd":10},
   "10": {"w":"befugt",           "pos":"VVPP",  "dep":"root",        "hd":-1},
   "11": {"w":".",                "pos":"\$.",    "dep":"punct",       "hd":10}}}
Reasoning: 0-6 = V1-advcl (dep=advcl, verb-first, no KOUS) -> 2.
           7 "so" = transition advmod -> 1. 8-11 = root clause -> 1.
Output: {"0":2, "1":2, "2":2, "3":2, "4":2, "5":2, "6":2, "7":1, "8":1, "9":1, "10":1, "11":1}

Example 2 --- Actor-as-condition + embedded \textit{soweit}-clause (discontinuous spans):

Sentence: "Bei jeder Übermittlung hat er , soweit dies möglich ist ,
           Informationen beizufügen ."
Input:
{"sent": "Bei jeder Übermittlung hat er , soweit dies möglich ist ,
          Informationen beizufügen .",
 "tokens": {
   "0":  {"w":"Bei",            "pos":"APPR",  "dep":"case",  "hd":2},
   "1":  {"w":"jeder",          "pos":"PIAT",  "dep":"det",   "hd":2},
   "2":  {"w":"Übermittlung",   "pos":"NN",    "dep":"obl",   "hd":3},
   "3":  {"w":"hat",            "pos":"VAFIN", "dep":"root",  "hd":-1},
   "4":  {"w":"er",             "pos":"PPER",  "dep":"nsubj", "hd":3},
   "5":  {"w":",",              "pos":"\$,",    "dep":"punct", "hd":8},
   "6":  {"w":"soweit",         "pos":"KOUS",  "dep":"mark",  "hd":8},
   "7":  {"w":"dies",           "pos":"PDS",   "dep":"nsubj", "hd":8},
   "8":  {"w":"möglich",        "pos":"ADJD",  "dep":"advcl", "hd":3},
   "9":  {"w":"ist",            "pos":"VAFIN", "dep":"cop",   "hd":8},
   "10": {"w":",",              "pos":"\$,",    "dep":"punct", "hd":12},
   "11": {"w":"Informationen",  "pos":"NN",    "dep":"obj",   "hd":12},
   "12": {"w":"beizufügen",     "pos":"VVIZU", "dep":"xcomp", "hd":3},
   "13": {"w":".",              "pos":"\$.",    "dep":"punct", "hd":3}}}
Reasoning: root=hat(3). er(4, nsubj) = actor-condition -> 2.
           soweit-clause (5-9, advcl+KOUS) -> 2. Rest -> 1 (discontinuous).
Output: {"0":1, "1":1, "2":1, "3":1, "4":2, "5":2, "6":2, "7":2, "8":2, "9":2, "10":1, "11":1, "12":1, "13":1}

Example 3 --- Demonstrative condition overrides actor rule:

Sentence: "Sie oder er hat in diesem Fall den Verantwortlichen zu informieren ."
Input:
{"sent": "Sie oder er hat in diesem Fall den Verantwortlichen zu informieren .",
 "tokens": {
   "0":  {"w":"Sie",               "pos":"PPER",   "dep":"nsubj",   "hd":3},
   "1":  {"w":"oder",              "pos":"KON",    "dep":"cc",      "hd":3},
   "2":  {"w":"er",                "pos":"PPER",   "dep":"nsubj",   "hd":3},
   "3":  {"w":"hat",               "pos":"VAFIN",  "dep":"root",    "hd":-1},
   "4":  {"w":"in",                "pos":"APPR",   "dep":"case",    "hd":6},
   "5":  {"w":"diesem",            "pos":"PDAT",   "dep":"det",     "hd":6},
   "6":  {"w":"Fall",              "pos":"NN",     "dep":"obl",     "hd":9},
   "7":  {"w":"den",               "pos":"ART",    "dep":"det",     "hd":8},
   "8":  {"w":"Verantwortlichen",  "pos":"NN",     "dep":"obl:arg", "hd":9},
   "9":  {"w":"zu",                "pos":"PTKZU",  "dep":"mark",    "hd":10},
   "10": {"w":"informieren",       "pos":"VVINF",  "dep":"xcomp",   "hd":3},
   "11": {"w":".",                 "pos":"\$.",     "dep":"punct",   "hd":3}}}
Reasoning: explicit condition "in diesem Fall" (4-6, PDAT obl) -> 2.
           Pronoun subjects (0-2) stay class 1 (explicit condition takes precedence).
           Rest -> 1.
Output: {"0":1, "1":1, "2":1, "3":1, "4":2, "5":2, "6":2, "7":1, "8":1, "9":1, "10":1, "11":1}

Example 4 --- No condition-consequence pair $\rightarrow$ all class 3:

Sentence: "Der Verantwortliche hat mit dem Bundesbeauftragten zusammenzuarbeiten ."
Input:
{"sent": "Der Verantwortliche hat mit dem Bundesbeauftragten zusammenzuarbeiten .",
 "tokens": {
   "0": {"w":"Der",                    "pos":"ART",   "dep":"det",   "hd":1},
   "1": {"w":"Verantwortliche",        "pos":"NN",    "dep":"nsubj", "hd":2},
   "2": {"w":"hat",                    "pos":"VAFIN", "dep":"root",  "hd":-1},
   "3": {"w":"mit",                    "pos":"APPR",  "dep":"case",  "hd":5},
   "4": {"w":"dem",                    "pos":"ART",   "dep":"det",   "hd":5},
   "5": {"w":"Bundesbeauftragten",     "pos":"NN",    "dep":"obl",   "hd":6},
   "6": {"w":"zusammenzuarbeiten",     "pos":"VVIZU", "dep":"xcomp", "hd":2},
   "7": {"w":".",                      "pos":"\$.",    "dep":"punct", "hd":2}}}
Reasoning: standing cooperation duty — no reconstructible Wenn/dann pair -> all 3.
Output: {"0":3, "1":3, "2":3, "3":3, "4":3, "5":3, "6":3, "7":3}

OUTPUT FORMAT:
Return ONLY a JSON object with key \texttt{results}: a list of dicts (one per sentence),
mapping token-index strings to class integers (1, 2, or 3).
Token count per dict MUST match the input. No explanations, no Markdown.

{"results": [{"0":2, "1":1, "2":1}, {"0":3, "1":3}]}

\textbf{User Prompt Template}

Classify these sentences token by token:

Sentence 0: <JSON object with keys \texttt{sent} (sentence string) and \texttt{tokens} (indexed token objects with fields \texttt{w}, \texttt{pos}, \texttt{dep}, \texttt{hd})>

Sentence 1: <same structure>

\ldots

Return results as pure JSON --- no explanations, no Markdown:
{"results": [{...sentence 0 labels...}, {...sentence 1 labels...}, ...]}

\section{Confusion Matrices}
\label{sec:confusion_matrices}
The confusion matrices show the predictions for each of the three classes. The relative values are grouped by the true label with the absolute values shown in parentheses. BiLSTM(-CRF) and BERT use the models with the best seeds.

\begin{figure*}
    \centering
    \includegraphics[width=1\linewidth]{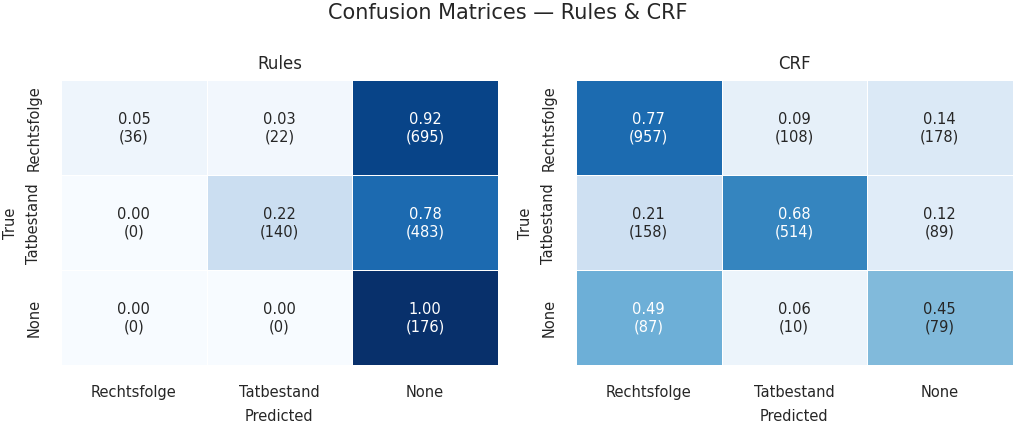}
    \caption{Confusion Matrices - Rules \& CRF}
    \label{fig:conf_rules_crf}
\end{figure*}

\begin{figure*}
    \centering
    \includegraphics[width=1\linewidth]{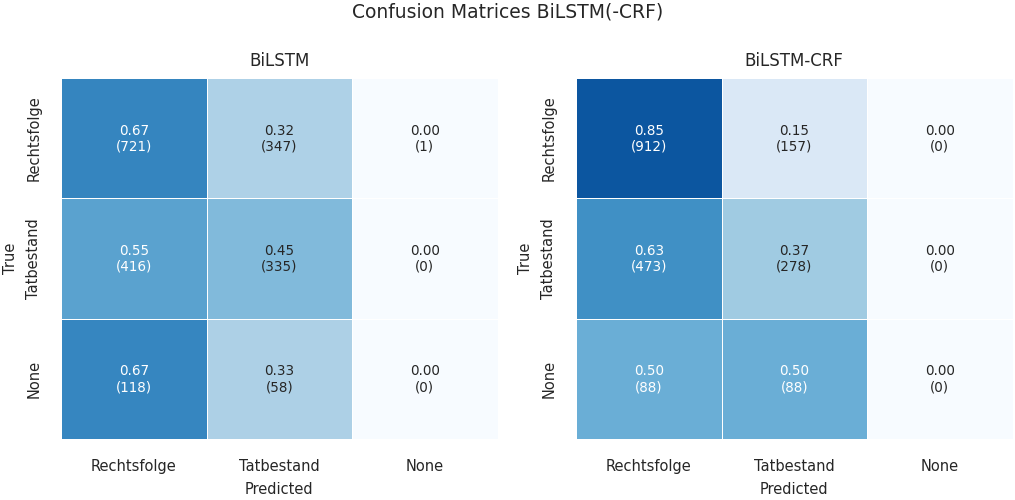}
    \caption{Confusion Matrices - BiLSTM \& BiLSTM-CRF}
    \label{fig:conf_bilstm}
\end{figure*}

\begin{figure*}
    \centering
    \includegraphics[width=1\linewidth]{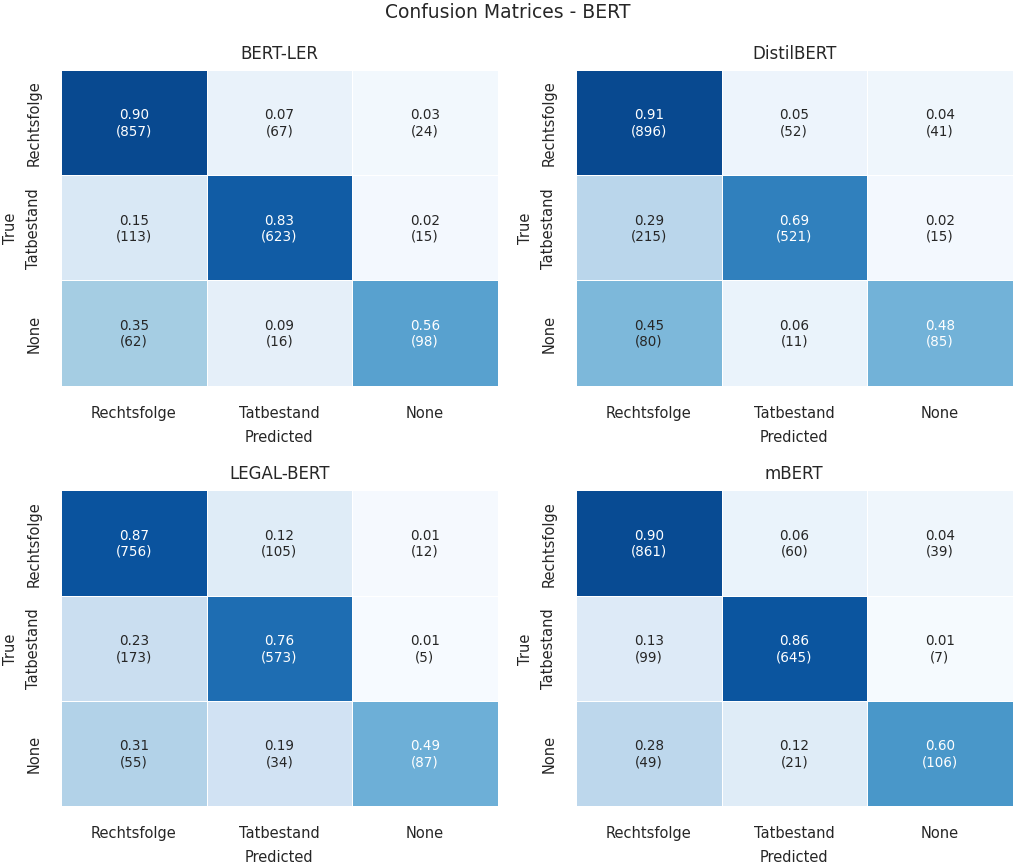}
    \caption{Confusion Matrices - BERT}
    \label{fig:conf_bert}
\end{figure*}

\begin{figure*}
    \centering
    \includegraphics[width=1\linewidth]{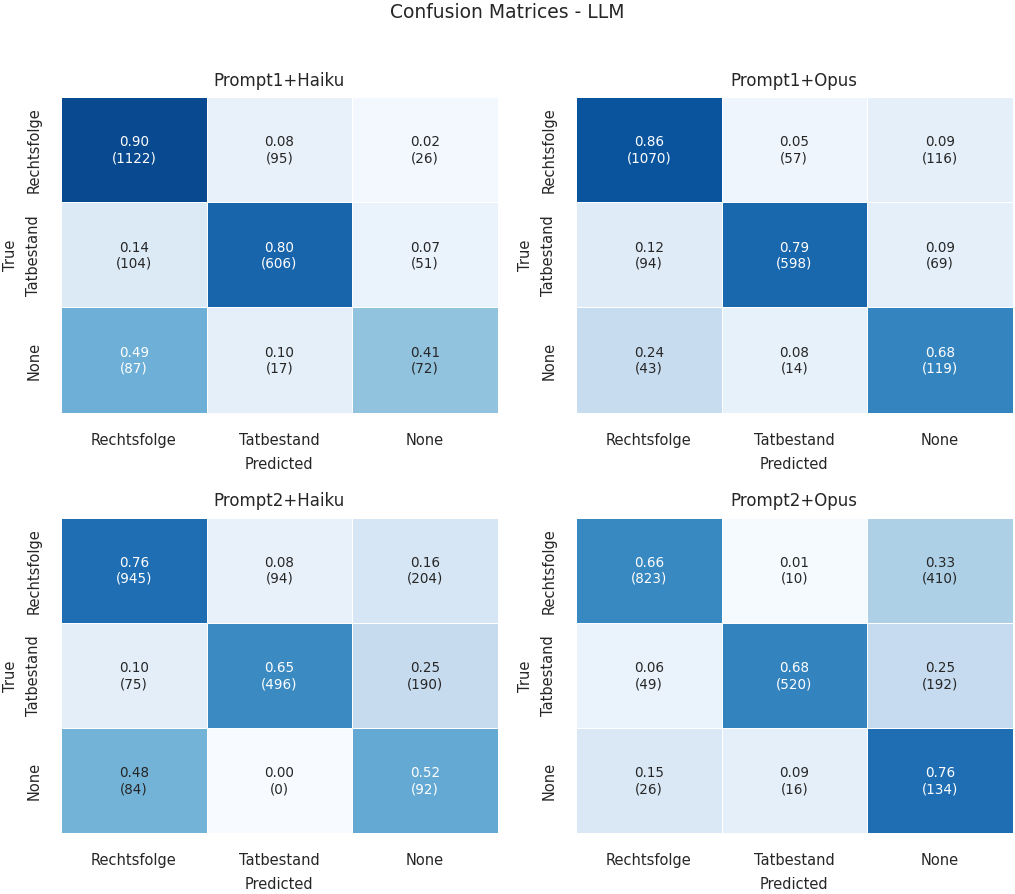}
    \caption{Confusion Matrices - LLM}
    \label{fig:conf_llm}
\end{figure*}

\begin{figure*}
    \centering
    \includegraphics[width=1\linewidth]{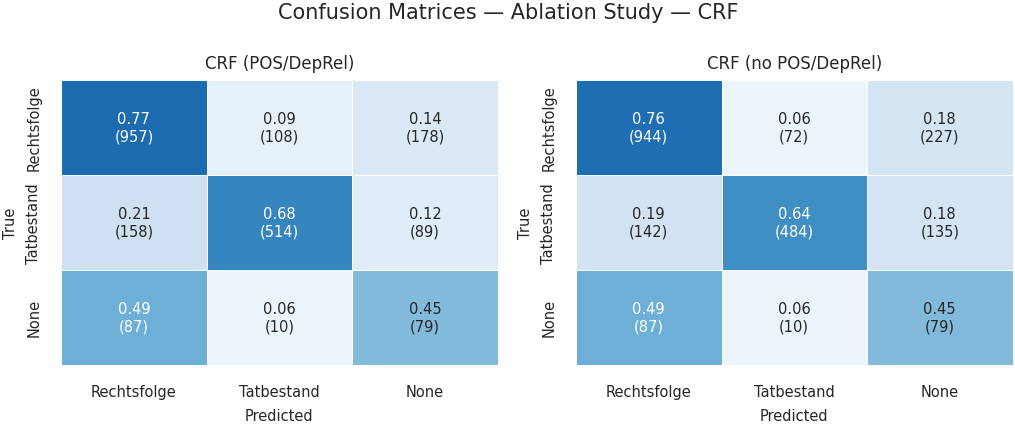}
    \caption{Confusion Matrices - Ablation study - CRF}
    \label{fig:conf_apl_crf}
\end{figure*}

\begin{figure*}
    \centering
    \includegraphics[width=1\linewidth]{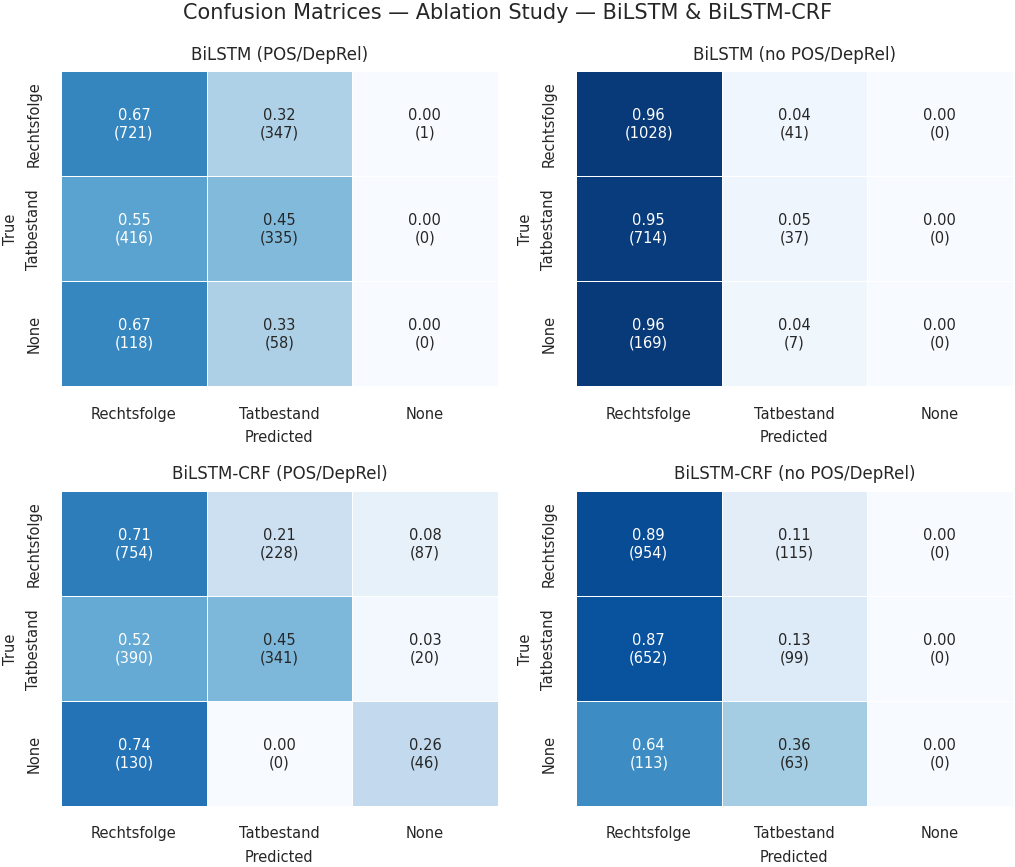}
    \caption{Confusion Matrices - Ablation study - BiLSTM \& BiLSTM-CRF}
    \label{fig:conf_apl_bilstm}
\end{figure*}

\begin{figure*}
    \centering
    \includegraphics[width=1\linewidth]{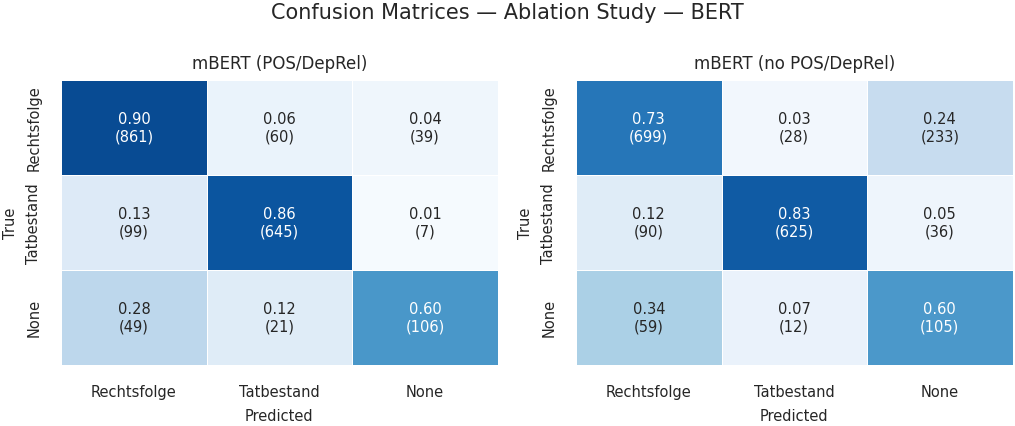}
    \caption{Confusion Matrices - Ablation study - BERT}
    \label{fig:conf_apl_bert}
\end{figure*}

\end{document}